\documentclass[conference]{IEEEtran}
\IEEEoverridecommandlockouts
\usepackage{cite}
\usepackage{amsmath,amssymb,amsfonts}
\usepackage{algorithmic}
\usepackage{graphicx}
\usepackage{textcomp}
\usepackage{xcolor}

\usepackage[symbol]{footmisc}

\def\BibTeX{{\rm B\kern-.05em{\sc i\kern-.025em b}\kern-.08em
    T\kern-.1667em\lower.7ex\hbox{E}\kern-.125emX}}
\begin{document}

\title{Split and Connect: A Universal Tracklet Booster for Multi-Object Tracking
}
\author{\IEEEauthorblockN{Gaoang Wang$^{1}$, Yizhou Wang$^{2}$, Renshu Gu$^{*,3}$, Weijie Hu$^{*,4}$, and Jenq-Neng Hwang$^{2}$}
\IEEEauthorblockA{$^1$\textit{Zhejiang University, China}\\
$^2$\textit{University of Washington, USA}\\
$^3$\textit{Hangzhou Dianzi University, China}\\
$^4$\textit{Guangdong University of Petrochemical Technology, China}\\
gaoangwang@intl.zju.edu.cn}}


\maketitle

\begin{abstract}
Multi-object tracking (MOT) is an essential task in the computer vision field. With the fast development of deep learning technology in recent years, MOT has achieved great improvement. However, some challenges still remain, such as sensitiveness to occlusion, instability under different lighting conditions, non-robustness to deformable objects, etc. To address such common challenges in most of the existing trackers, in this paper, a tracklet booster algorithm is proposed, which can be built upon any other tracker. The motivation is simple and straightforward: split tracklets on potential ID-switch positions and then connect multiple tracklets into one if they are from the same object. In other words, the tracklet booster consists of two parts, i.e., Splitter and Connector. First, an architecture with stacked temporal dilated convolution blocks is employed for the splitting position prediction via label smoothing strategy with adaptive Gaussian kernels. Then, a multi-head self-attention based encoder is exploited for the tracklet embedding, which is further used to connect tracklets into larger groups. We conduct sufficient experiments on MOT17 and MOT20 benchmark datasets, which demonstrates promising results. Combined with the proposed tracklet booster, existing trackers usually can achieve large improvements on the IDF1 score, which shows the effectiveness of the proposed method. 
\end{abstract}


\footnotetext[1]{Corresponding author.}

\section{Introduction}
Multi-object tracking (MOT) has drawn great attention in recent years. This technique is critically needed in many tasks, such as traffic flow analysis \cite{tang2018single,tang2019cityflow,wang2019anomaly,hsu2019multi}, human behavior prediction and pose estimation \cite{gu2019efficient,gu2020exploring,gu2019multi,jagadeesh2018human,jalal2019multi}, autonomous driving assistance \cite{kimeagermot,chaabane2021deft,hu2019joint} and even for underwater animal abundance estimation \cite{wang2016closed,chuang2016underwater,dawkins2017open}. Recent years have seen the emergence of a variety of tracking algorithms, from graph clustering methods \cite{tang2017multiple,milan2016multi,tang2018single,keuper2016multi,kumar2014multiple} to graph neural networks \cite{wang2020joint,braso2020learning,shan2020fgagt,weng2020gnn3dmot} that aggregate information across frames and objects, from tracking-by-detection paradigm to joint detection and tracking \cite{zhou2020tracking,peng2020chained,wang2020joint,pang2020tubetk,tracktor_2019_ICCV,wang2019towards} to improve the detection performance with multiple frames, from Kalman filtering \cite{kim2014data} to recurrent neural networks (RNN) \cite{milan2017online} and long-short term memory (LSTM) \cite{lu2017online} to boost association performance with the motion clue. However, due to the noisy visual object detection and occlusion, tracking multiple objects over a long time is yet very challenging. 

\begin{figure}[t]
\begin{center}
\includegraphics[width=1.0\linewidth]{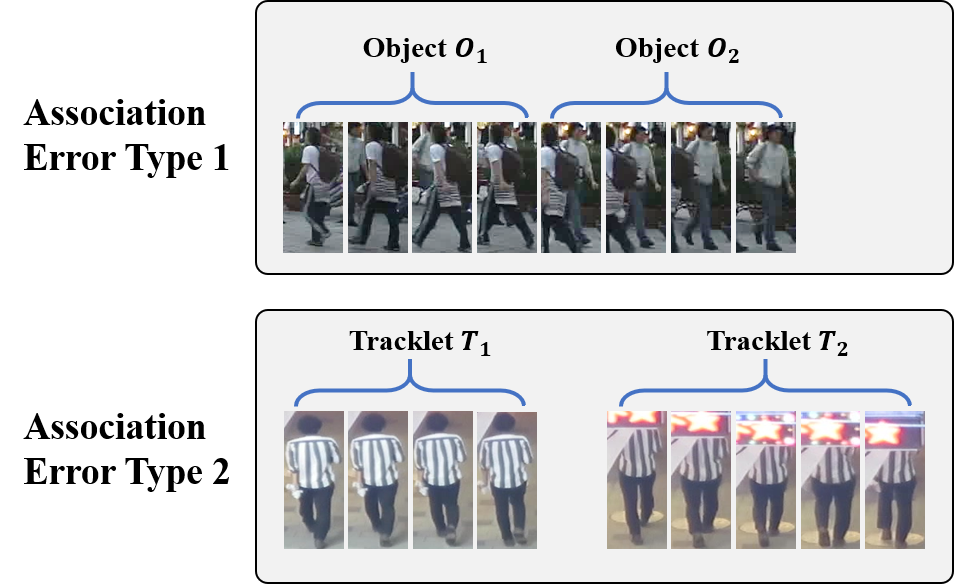}
\end{center}
   \caption{Demonstration of association errors. The first error type shows multiple objects are associated to the same tracklet. The second error shows the same object are assigned to different tracklets.}
\label{fig:error}
\end{figure}

Typically, the tracking error comes from two parts: object detection and temporal association. To measure the performance of trackers, three main evaluation metrics, i.e., multi-object tracking accuracy (MOTA) \cite{milan2016mot16}, IDF1 score \cite{ristani2016performance} and higher order tracking accuracy (HOTA) \cite{luiten2020hota}, are widely used in MOT field. As demonstrated in \cite{luiten2020hota}, MOTA emphasizes detection accuracy while IDF1 focuses more on association measurement. As shown in Figure~\ref{fig:error}, generally, the association errors can be concluded into two categories: 1) different objects are associated to the same tracklet, 2) tracklets from the same object are assigned to different IDs. A good tracker should reduce these two types of errors as much as possible in the association task. Due to missing detection, change of lighting condition, camera movement, occlusion and object deformation, the association is one of the major challenges of almost all existing trackers.

\begin{figure*}[!t]
\begin{center}
\includegraphics[width=1.0\linewidth]{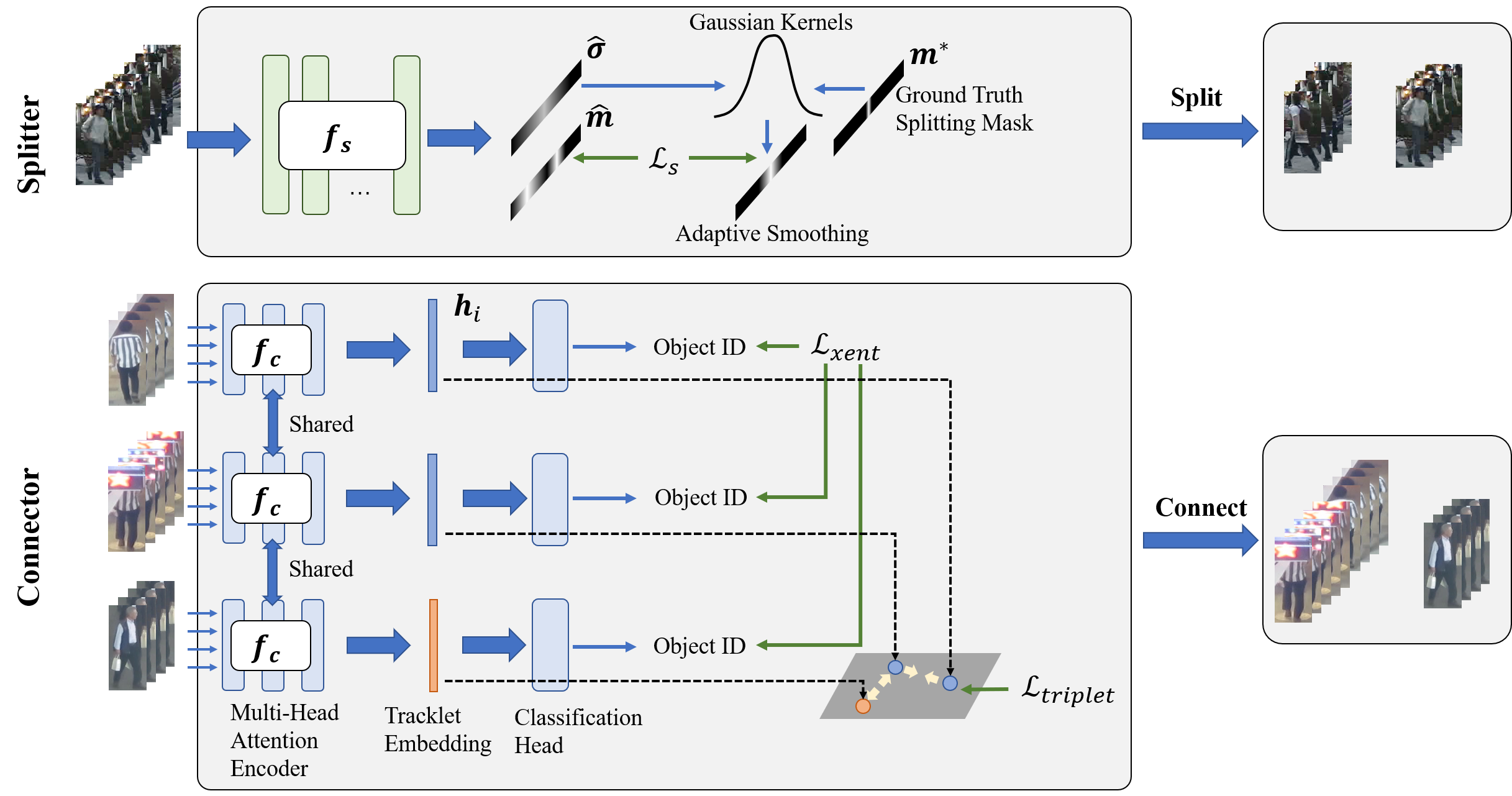}
\end{center}
   \caption{The flowchart of the proposed tracklet booster. The top part and bottom part show the process of Splitter and Connector, respectively. $f_s$ and $f_c$ represent the dilated temporal convolution model and multi-head self-attention based encoder used in Splitter and Connector, respectively. $\hat{\boldsymbol{\sigma}}$ is the predicted standard deviation for adaptive smoothing; $\hat{\boldsymbol{m}}$ is the predicted splitting position mask; $\boldsymbol{m^*}$ is the ground truth splitting mask; $\mathcal{L}_s$, $\mathcal{L}_{xent}$ and $\mathcal{L}_{triplet}$ are mean squared error, cross-entropy loss and triplet loss, respectively.}
\label{fig:flowchart}
\end{figure*}

In this paper, we propose a novel method, tracklet booster, that directly targets the two types of association error to boost the performance on IDF1 score, without tackling much for the detection error. Also, it can be efficiently plugged into any existing trackers. The tracklet booster has two main modules, Splitter and Connector. Splitter aims at finding potential ID-switch positions of a tracklet and split the tracklet into small pieces at the detected ID-switch positions. Splitter is designed with stacked dilated temporal convolution blocks to measure the temporal consistency of a tracklet. An adaptive label smoothing strategy with Gaussian kernels is proposed to improve the stability of the model training. With Splitter, the first type of error that multiple objects are assigned with the same ID can be largely reduced. On the other hand, Connector is introduced to address the second type of association error that the same identity is assigned to multiple tracklets. Connector aims at distinguishing different objects and merging multiple tracklets into one if they are from the same object. Specifically, Connector is modeled as a tracklet embedding network. Tracklets with small embedding distances are grouped and associated with the same tracking ID. Inspired by the transformer \cite{vaswani2017attention}, we use the multi-head self-attention mechanism to learn the tracklet embedding. The framework of the tracklet booster is shown in Figure~\ref{fig:flowchart}.

We summarize our contributions as follows:
\begin{itemize}
\item We propose a novel tracklet boosting model, consisting of a Splitter and a Connector, to directly address the temporal association errors that exist in almost all trackers in the MOT field. Besides, the proposed tracklet booster can be integrated with any existing trackers to significantly improve their tracking performance.
\item A novel adaptive label smoothing strategy with Gaussian kernels is proposed in Splitter to predict the potential ID-switch positions within the given tracklet.
\item Multi-head self-attention based encoder is employed for tracklet embedding, which serves as the main module of the proposed Connector.
\item We conduct experiments on MOT17 and MOT20 benchmark datasets and prove the generality, effectiveness and robustness of our tracklet boosting method.
\end{itemize}

\section{Related Work}

\subsection{Tracking with Graph Models}

Graph models \cite{tang2017multiple,milan2016multi,tang2018single,keuper2016multi,kumar2014multiple,choi2015near,tang2015subgraph,wen2014multiple,wang2019exploit,wang2017learning,lenz2015followme,milan2013continuous,andriyenko2012discrete,hornakova2020lifted} are widely used in MOT for temporal association. The association is traditionally solved by optimizing the total cost or energy function. 
For example, \cite{milan2016multi} formulates MOT as a minimization of a continuous energy function. \cite{tang2017multiple} proposes a novel graph-based formulation that links and clusters person hypotheses over time by solving an instance of a minimum cost lifted multi-cut problem. \cite{wang2017learning} proposes an end-to-end framework for learning parameters of min-cost flow MOT problem with quadratic trajectory interactions including suppression of overlapping tracks and contextual cues about the co-occurrence of different objects. \cite{hornakova2020lifted} proposes an extension to the disjoint paths problem in which additional lifted edges are introduced to provide path connectivity priors. \cite{lenz2015followme} proposes an efficient online min-cost flow tracking algorithm with bounded memory and computation. 
In addition to the single-view tracking, graph based methods are also explored in multi-view tracking tasks \cite{xu2016multi,xu2017cross,han2020cvmht}, where the multi-view tracking is formulated as a graph clustering problem. 
Usually, detections or tracklets are adopted as graph nodes. Then, the similarities among nodes are measured on the connected edges. For detection-based graphs, the temporal information is not well utilized and usually comes with a remarkably high dimensional affinity matrix with a heavy computational cost. The tracklet-based graph, on the other hand, is much more efficient and also incorporates global temporal information. However, the conventional graph models, based on optimization, usually suffer from empirically setting of hand-crafted features. Moreover, representative embeddings are not well-explored for the temporal association.

Since the graph neural networks (GNN) show great power recently, many approaches \cite{wang2020joint,braso2020learning,shan2020fgagt,weng2020gnn3dmot,he2021learnable} adopt GNN for the association, rather than using conventional graph models based on optimization. Specifically, \cite{braso2020learning} exploits the classical network flow formulation of MOT to define a fully differentiable framework based on message passing networks. \cite{shan2020fgagt} presents an adaptive graph neural network to fuse locations, appearance, and historical information for MOT. \cite{weng2020gnn3dmot} proposes a novel feature interaction mechanism based on the GNN to learn the interaction among objects. \cite{he2021learnable} presents a novel learnable graph matching method to address association issues in MOT. \cite{dai2021learning} proposes a novel proposal-based learnable framework, which models MOT as a proposal generation, proposal scoring and trajectory inference paradigm on an affinity graph. The existing methods show the effectiveness of employing GNN to MOT. 
Moreover, GNN also shows power in other related vision tasks, such as human action recognition \cite{guo2018neural}, visual question answering \cite{narasimhan2018out} and single object tracking \cite{gao2019graph}. 

\subsection{Joint Detection and Tracking}
More recently, joint detection and tracking based methods have drawn great attention \cite{zhou2020tracking,peng2020chained,wang2020joint,pang2020tubetk,tracktor_2019_ICCV,wang2019towards,pang2020quasi}. Usually, such trackers take sequential adjacent frames as input. Features are aggregated in different frames, and bounding box regression is conducted with temporal information. For example, Tracktor \cite{tracktor_2019_ICCV} applies bounding box regression to refine the box position in the next frame to form the track; JDE \cite{wang2019towards} incorporates the appearance embedding model into a single-shot detector that can simultaneously
output detections and the corresponding embeddings; while CenterTrack \cite{zhou2020tracking} applies a detection model to a pair of images and detections from the prior frame. Besides that, CTracker \cite{peng2020chained} constructs tracklets by chaining paired boxes in every two frames. TubeTK \cite{pang2020tubetk} directly predicts a box tube as a tracklet in an offline manner. \cite{pang2020quasi} proposes a quasi-dense similarity learning approach that densely samples hundreds of region proposals on a pair of images for contrastive learning. However, due to the heavy computational cost, the networks can only take a very limited number of frames as input. However, such methods usually suffer long-time occlusions, which further results in temporal association error.

\subsection{Tracking with Visual Transformers}
Recently, due to the non-local attention mechanism, transformers \cite{vaswani2017attention} show great success in many visual tasks, such as image classification \cite{dosovitskiy2020image}, object detection \cite{carion2020end,zhu2020deformable}, 3D human pose estimation \cite{zheng20213d} and low-level image processing \cite{chen2020pre}. Moreover, several methods with visual transformers \cite{sun2020transtrack,meinhardt2021trackformer,chu2021spatial} are explored in the MOT field. For example, \cite{sun2020transtrack} proposes a baseline tracker via transformer, which takes advantage of the query-key mechanism and introduces a set of learned object queries into the pipeline to enable detecting new-coming objects. \cite{meinhardt2021trackformer} extends the DETR object detector \cite{carion2020end} and achieves a seamless data association between frames in a new tracking-by-attention paradigm by encoder-decoder and self-attention mechanisms. \cite{chu2021spatial} leverages graph transformers to efficiently model the spatial and temporal interactions among the objects. However, transformer-based trackers usually require pre-training on large-scale datasets.

\section{Method}

The motivation of the tracklet booster is simple yet effective. We take the tracklets from any tracklet generators \cite{wang2016closed,zhang2017multi,wang2019exploit} or preliminary tracking results as the input. Due to matching errors, a tracklet may contain multiple object IDs. To clean the IDs within the tracklet, Splitter is proposed to split tracklets into small pieces on the potential ID-switch positions to ensure split tracklets have purer IDs as much as possible. Next, the split tracklets are sent into Connector to learn representative embeddings. Finally, the tracklets with the similar embeddings are grouped to form clusters, i.e., to generate the final entire track. The framework of the tracklet booster is shown in Figure~\ref{fig:flowchart}. The details of Splitter and Connector are demonstrated in the following sub-sections.

\subsection{Splitter}

In this sub-section, we demonstrate the proposed Splitter, which is designed to predict the potential ID-switch positions and split the tracklets. Denote the input tracklets as $\boldsymbol{x}$ with the dimension $K\times T$, where $K$ is the feature dimension, and $T$ is the temporal length of tracklet. Denote the ID-switch position mask as $\boldsymbol{m}$ with the dimension $T-1$, where $\boldsymbol{m}$ is set as follows,
\begin{equation}
    \boldsymbol{m}_{t}=
    \begin{cases}
    1, \ \text{if} \ \text{ID}_{t} \neq \text{ID}_{t+1}; \\
    0, \ \text{otherwise}.
    \end{cases}
\label{eq:m}
\end{equation}
The goal of Splitter is to estimate the ID-switch position mask given the input tracklets $\boldsymbol{x}$, i.e.,
\begin{equation}
    \boldsymbol{\hat m} = f_{s}(\boldsymbol{x}),
\label{eq:mask}
\end{equation}
where $f_s(\cdot)$ is the proposed Splitter. 

\begin{figure}[t]
\begin{center}
\includegraphics[width=1.0\linewidth]{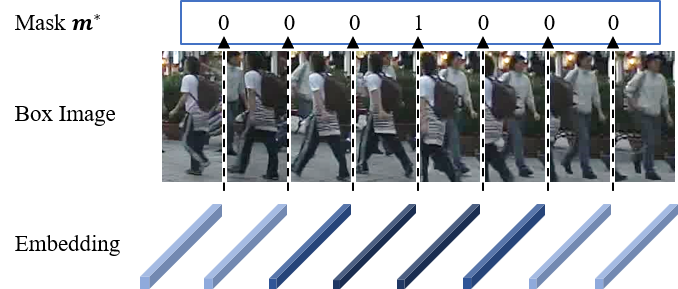}
\end{center}
   \caption{Example of the transition of an ID switch. Usually, the embeddings around the ID switch positions change gradually. Thus, the hard label of the ground truth splitting position mask $\boldsymbol{m}^*$ can cause instability in the training.}
\label{fig:split_mask}
\end{figure}

To learn Splitter, a loss function, i.e., $\mathcal{L}_s=l(\boldsymbol{\hat m}, \boldsymbol{m}^*)$, needs to be defined between the predicted ID-switch position mask $\boldsymbol{\hat m}$ and the ground truth ID-switch position mask $\boldsymbol{m}^*$. As shown in Figure~\ref{fig:split_mask}, the object embeddings in adjacent frames usually change gradually rather than change abruptly, even in the ID-switch positions. However, the ground truth mask is hard labeled with zeros and ones. Such inconsistency between labels and features will harm the stability of the model training with the commonly used mean squared error, i.e., $\mathcal{L}_s=||\boldsymbol{\hat m}-\boldsymbol{m}^*||^2$. As a result, it would be more appropriate to use soft labels rather than hard labels in the loss function design.

To incorporate the soft label, we adopt Gaussian smoothing in the mean squared error as the loss function as follows,
\begin{equation}
    \mathcal{L}_s = \sum_t \left(\boldsymbol{\hat m}_t-\min\left(\sum_\tau \boldsymbol{m}^*_{\tau}\exp \left(-{\frac{(\tau-t)^2}{\sigma^2}}\right), 1 \right)\right)^2,
\label{eq:loss}
\end{equation}
where $t$ and $\tau$ are the frame index for the predicted mask $\boldsymbol{\hat m}$ and ground truth mask $\boldsymbol{m}^*$, respectively; $\sigma$ is the standard deviation that controls the smoothness of the label. The ground truth mask $\boldsymbol{m}^*$ is used as an indicator for the summation of Gaussian kernels. We also use $\min(\cdot, 1)$ to constrain the smoothed mask labels in the range $[0, 1]$.

In different scenarios, the gradual change of embeddings when the ID switch happens can last from a few frames to tens of frames. In other words, $\sigma$ should be dependent on the duration of the ID-switch transition. Setting a fixed $\sigma$ in the model would derive a sub-optimal solution. To deal with such an issue, we propose an adaptive Gaussian smoothing strategy. Along with the prediction of ID-switch position mask $\boldsymbol{\hat m}$, we also predict a time and tracklet dependent $\boldsymbol{\hat \sigma}$ simultaneously via Splitter as follows,
\begin{equation}
    [\boldsymbol{\hat m}, \boldsymbol{\hat \sigma}] = f_{s}(\boldsymbol{x}),
\label{eq:mask_sigma}
\end{equation}
where $\boldsymbol{\hat \sigma}$ has the same dimension as $\boldsymbol{\hat m}$. Based on the time dependent $\boldsymbol{\hat \sigma}$, we modify the loss function as follows,
\begin{equation}
    \mathcal{L}_s = \sum_t \left(\boldsymbol{\hat m}_t-\min \left(\sum_\tau \boldsymbol{m}^*_{\tau}\exp\left(-{\frac{(\tau-t)^2}{\boldsymbol{\tilde \sigma}_{\tau}^2}}\right), 1 \right)\right)^2.
\label{eq:loss_sigma}
\end{equation}
where we set $\boldsymbol{\tilde \sigma}=\max(\min(\boldsymbol{\hat \sigma}, 10), \epsilon)$ to avoid irregular predictions. Typically, $\epsilon$ is set to be a small positive scalar, e.g., 0.001. With adaptive soft labels, Splitter model can predict ID-switch positions with different variant transitions of ID switches.

\begin{figure}[t]
\begin{center}
\includegraphics[width=1.0\linewidth]{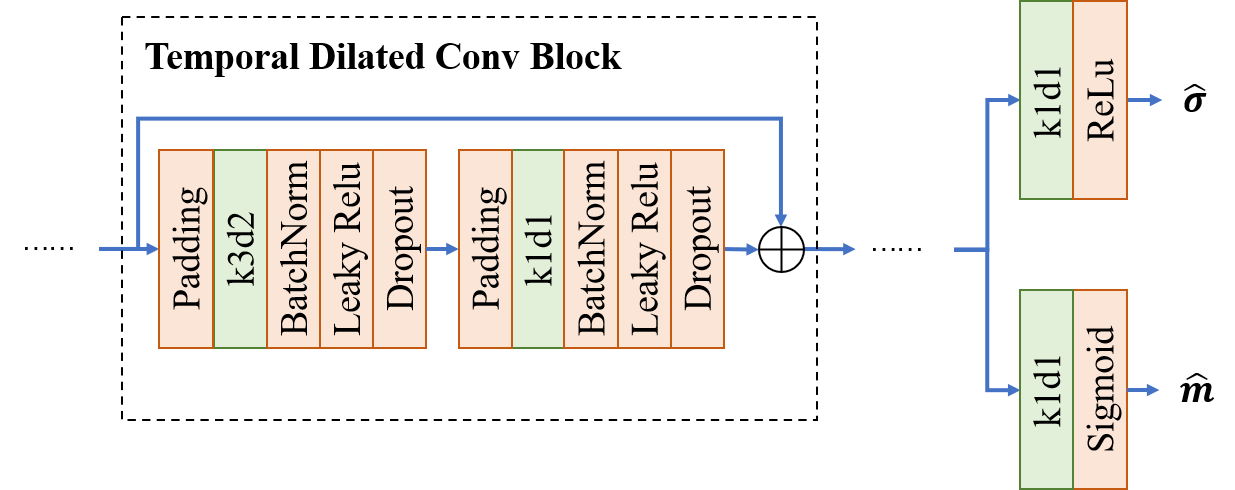}
\end{center}
   \caption{The architecture of the Splitter model. The box with dashed lines represents each temporal dilated convolution block. ``k3d2'' means the temporal convolution with kernel size 3 and dilation rate 2. ``$\oplus$'' represents the summation operation.}
\label{fig:splitter}
\end{figure}

The architecture of Splitter is designed as follows. Commonly used stacked dilated temporal convolution and pointwise convolution blocks are employed as the backbone for the feature extraction. Then two fully connected layers are used for ID-switch position mask and standard deviation prediction, respectively. We also add a skip connection for each intermediate block. In total, we stack 24 blocks. As a result, the receptive field is large enough to capture the long-term temporal patterns to detect the splitting positions. The architecture of Splitter is shown in Figure~\ref{fig:splitter}.

\subsection{Connector}

In this sub-section, we demonstrate the proposed Connector, which is designed to connect multiple tracklets if they are from the same object. We formulate the tracklet connection as a tracklet embedding problem as follows,
\begin{equation}
    \boldsymbol{h} = f_{c}(\boldsymbol{x}),
\label{eq:emb}
\end{equation}
where $f_{c}$ is the proposed Connector model, $\boldsymbol{h}$ is the tracklet embedding with $\boldsymbol{h} \in \mathbb{R}^{D}$. After embedding is learned, tracklets from the same object should have a smaller embedding distance, while the tracklets from distinct objects should have a larger distance. 

\begin{figure}[t]
\begin{center}
\includegraphics[width=1.0\linewidth]{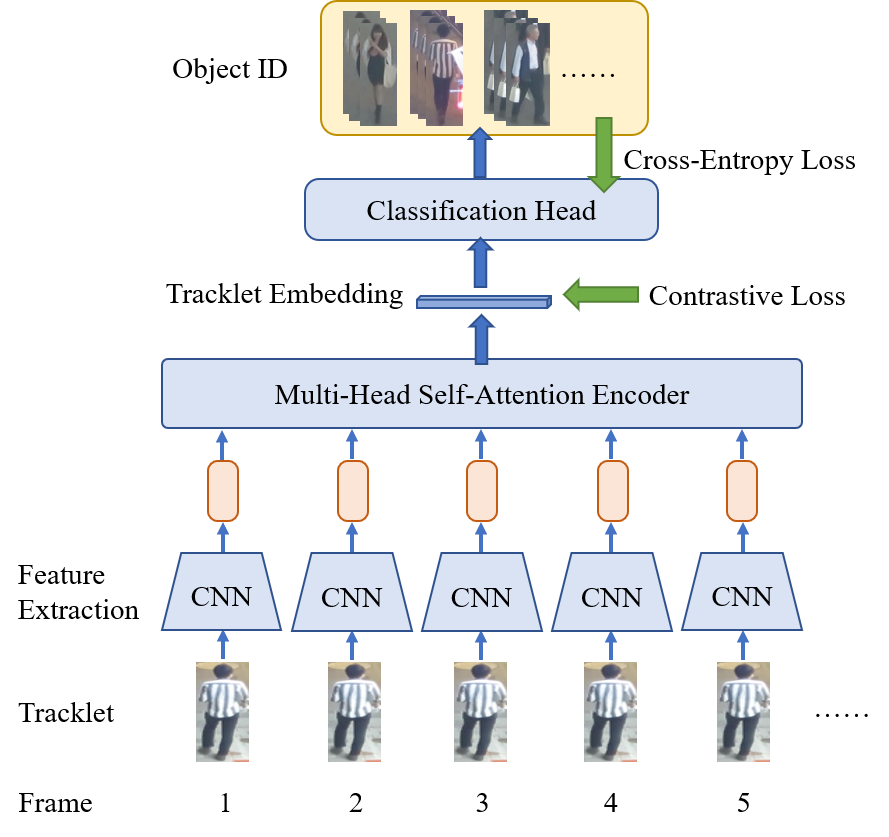}
\end{center}
   \caption{Connector model. Input tracklets are first sent to a pre-trained CNN for feature extraction. A multi-head self-attention based encoder is employed with the combination of cross-entropy loss and contrastive loss for tracklet embedding.}
\label{fig:connector}
\end{figure}

Due to sensitivity in occlusion, the difference in lighting condition, deformation in object pose, features from tracklets may have much difference even though they are from the same object. To address such issues, self-attention is an appropriate strategy for tracklet embedding. Inspired from the transformer \cite{vaswani2017attention}, a multi-head self-attention mechanism is employed for tracklet embedding in the proposed Connector model. 
The embedding framework is shown in Figure~\ref{fig:connector}.


Specifically, given the tracklet appearance features $\boldsymbol{x} \in \mathbb{R}^{K\times T}$, the self-attention (SA) is defined as follows,
\begin{equation}
\label{eq:self_at}
\begin{aligned}
    & [\boldsymbol{q}, \boldsymbol{k}, \boldsymbol{v}] = \boldsymbol{x}^T\boldsymbol{W}, \\
    & \boldsymbol{A} = \text{softmax}\left(\boldsymbol{q}\boldsymbol{k}^T/\sqrt{D_h}\right),\\
    & \text{SA}(\boldsymbol{x}) = \boldsymbol{A}\boldsymbol{v},
\end{aligned}
\end{equation}
where $\boldsymbol{W} \in \mathbb{R}^{D\times 3D_h}$ is the transformation that converts the input tracklet to query, key and value; $\boldsymbol{A} \in \mathbb{R}^{T\times T}$ stores the attention weights; $\text{SA}(\boldsymbol{x})$ represents the self-attention given the input $\boldsymbol{x}$. For the first input layer, we set $D=K$. Then we define the multi-head self-attention with the concatenation of $k$ SA as follows,
\begin{equation}
    \text{MSA}(\boldsymbol{x}) = [\text{SA}_1(\boldsymbol{x}); \text{SA}_2(\boldsymbol{x});...;\text{SA}_k(\boldsymbol{x})]\boldsymbol{W}^O,
\label{eq:msa}
\end{equation}
where $\boldsymbol{W}^O \in \mathbb{R}^{kD_h\times D}$ aggregates the information from multi-heads and transforms it back to the original dimension $D$. After the multi-head self-attention layer, a feed-forward layer is further applied for encoding.


After the final layer of the encoder, we also stack a classification head for learning the embedding. The classification head includes a global average pooling (GAP) layer, $l_2$ normalization and one fully connected layer, i.e.,
\begin{equation}
\label{eq:classification}
\begin{aligned}
    & \boldsymbol{h} = \text{norm}_{l_2}\left(\text{GAP}(\boldsymbol{z}_L)\right), \\
    & \boldsymbol{p} = \boldsymbol{W}_c\boldsymbol{h},
\end{aligned}
\end{equation}
where $\boldsymbol{z}_L$ is the output of the last encoding layer, $\boldsymbol{h}$ is the final tracklet embedding, $\boldsymbol{W}_c \in \mathbb{R}^{D \times C}$ is the weight of the fully connected layer, $\boldsymbol{p}$ is the predicted logits for $C$ classes. The purpose of GAP is to pool the features along the temporal dimension from $D \times T$ to $D$. cross-entropy loss $\mathcal{L}_{xent}$ and triplet loss $\mathcal{L}_{triplet}$ are used in the training as follows,
\begin{equation}
\label{eq:loss_connector}
\begin{aligned}
    & \mathcal{L}_{xent} = \sum_{i=1}^N -\log\left(\frac{\exp\left(\boldsymbol{p}_i^c\right)}{\sum_{j=1}^C\exp\left(\boldsymbol{p}_i^j\right)}\right), \\
    & \mathcal{L}_{triplet} = \sum_{i=1}^N [||\boldsymbol{h}_i^a-\boldsymbol{h}_i^p||_2-||\boldsymbol{h}_i^a-\boldsymbol{h}_i^n||_2+\alpha]_+,\\
    & \mathcal{L}_{total} = \mathcal{L}_{xent}+\lambda\mathcal{L}_{triplet},
\end{aligned}
\end{equation}
where $\boldsymbol{h}_i^a$, $\boldsymbol{h}_i^p$ and $\boldsymbol{h}_i^n$ are embeddings of the anchor sample, the positive sample and the negative sample in the mini-batch, $\alpha$ is a pre-defined distance margin, and $\lambda$ is the trade-off between two loss terms. 

For the architecture of Connector, we stack 6 encoding layers. 4-head self-attention is employed in each encoding layer. We set $D=512$ as the intermediate channel dimension.

\subsection{Tracking}
\label{sec:inference}
In the inference stage, each of the initial tracklets is first sent to the Splitter model. Based on the predicted ID-switch position mask $\boldsymbol{\hat m}$, local maximum peaks are picked and the peak values are compared with a pre-defined splitting threshold $\delta_s$. The positions with peak values greater than $\delta_s$ are treated as splitting positions. Then tracklets are split at the predicted splitting positions. 

To group tracklets into tracks with the trained Connector, a tracklet graph $\mathcal{G}(V, E)$ is first built based on the split tracklets, where $V$ and $E$ are the vertex set and edge set, respectively. Specifically, each tracklet is treated as a vertex in set $V$, while $E$ is a finite set in which every element $e \in E$ represents an edge between a pair of two tracklets $u,w \in V$ that are not far away in the time domain, yet no temporal overlap with each other, i.e.,
\begin{equation}
    \min_{t_u\in T(u), t_w \in T(w)} \lvert t_u - t_w \rvert \leq \delta_t \text{ and } 
    T_f(u) \cap T_f(w)=\emptyset,
\end{equation}
where $T_f(u)$ is the set of frame indices of the tracklet $u$. Based on the built tracklet graph, each tracklet is embedded via the encoder of the proposed Connector. A pairwise Euclidean distance on each element $e \in E$ is measured. With a sufficiently good embedding framework, a simple bottom-up greedy matching algorithm is employed to group tracklets if two tracklets satisfy 
\begin{equation}
    ||\boldsymbol{h}_i-\boldsymbol{h}_j||_2<\delta_c,
\end{equation}
where $\delta_c$ is a pre-defined connection threshold. The setting of the hyper-parameters is described in Section~\ref{sec:inf_detail}.

The time complexity for Splitter and Connector is as follows. Denote the number of tracklets as $M$, and the video length as $L$. Since Splitter is conducted in a temporal window manner, the time complexity in inference is $O(ML)$. For Connector, the tracklet embedding takes about $O(ML)$, while the graph clustering with the greedy method takes about $O(M^2L)$. As a result, the complexity of the entire model is roughly $O(M^2L)$. 

\section{Experiments}

\begin{table*}[!t]
\begin{center}
\caption{Result on MOT17 testing set. For each pair of comparisons, the first row shows the original state-of-the-art (SOTA) method and the second row shows the corresponding method with the proposed TBooster (short for ``tracklet booster''). The number in ``()'' besides the IDF1 score is the improvement after TBooster applied.}
\label{tab:MOT17}
\begin{tabular}{l|cccccccc}
\hline
Method (MOT17) & IDF1 (\%) $\uparrow$ & MOTA (\%) $\uparrow$ & MOTP (\%) $\uparrow$ & MT (\%) $\uparrow$ & ML (\%) $\downarrow$ & IDS $\downarrow$ & FRAG $\downarrow$ & FPS $\uparrow$\\
\hline
\hline
IOU \cite{1517Bochinski2017} & 39.4 & 45.5 & 76.9 & 369 & 953 & 5,988 & 7,404 & 1,522.9\\
\textbf{TBooster+IOU} & \textbf{45.1} \color{red}{(+5.7)} & 45.8 & 76.8 & 369 & 953 & 4,189 & 7,430 & 7.7\\
\hline
Tracktor\_v2 \cite{tracktor_2019_ICCV} & 55.1 & 56.3 & 78.8 & 498 & 831 & 1,987 & 3,763 & 1.5\\
\textbf{TBooster+Tracktor\_v2} & \textbf{59.2} \color{red}{(+4.1)} & 56.4 & 78.9 & 498 & 831 & 1,785 & 3,750 & 1.3\\
\hline
MPNTrack \cite{braso2020learning} & 61.7 & 58.8 & 78.6 & 679 & 788 & 1,185 & 2,265 & 6.5\\
\textbf{TBooster+MPNTrack} & \textbf{62.3} \color{red}{(+0.6)} & 58.9 & 78.7 & 682 & 790 & 1,198 & 2,209 & 4.2\\
\hline
CenterTrack \cite{zhou2020tracking} & 59.6 & 61.5 & 78.9 & 621 & 752 & 2,583 & 4,965 & 17.0\\
\textbf{TBooster+CenterTrack} & \textbf{63.3} \color{red}{(+3.7)} & 61.5 & 78.8 & 622 & 754 & 2,470 & 5,079 & 6.9\\
\hline
\end{tabular}
\end{center}
\end{table*}

\begin{table*}[!t]
\begin{center}
\caption{Result on MOT20 testing set. For each pair of comparisons, the first row shows the original state-of-the-art (SOTA) method and the second row shows the corresponding method with the proposed TBooster (short for ``tracklet booster''). The number in ``()'' besides the IDF1 score is the improvement after TBooster applied.}
\label{tab:MOT20}
\begin{tabular}{l|cccccccc}
\hline
Method (MOT20) & IDF1 (\%) $\uparrow$ & MOTA (\%) $\uparrow$ & MOTP (\%) $\uparrow$ & MT (\%) $\uparrow$ & ML (\%) $\downarrow$ & IDS $\downarrow$ & FRAG $\downarrow$ & FPS $\uparrow$\\
\hline
\hline
Tracktor \cite{tracktor_2019_ICCV} & 52.7 & 52.6 & 79.9 & 365 & 331 & 1,648 & 4,374 & 1.2\\
\textbf{TBooster+Tracktor} & \textbf{53.3} \color{red}{(+0.6)} & 52.6 & 79.8 & 365 & 329 & 1,734 & 4,389 & 0.6\\
\hline
UnsupTrack \cite{karthik2020simple} & 50.6 & 53.6 & 80.1 & 376 & 311 & 2,178 & 4,335 & 1.3\\
\textbf{TBooster+UnsupTrack} & \textbf{54.3} \color{red}{(+3.7)} & 53.7 & 80.1 & 374 & 313 & 1,771 & 4,322 & 0.6\\
\hline
MOT20\_TBC \cite{ren2020tracking} & 50.1 & 54.5 & 77.3 & 415 & 245 & 2,449 & 2,580 & 5.6\\
\textbf{TBooster+MOT20\_TBC} & \textbf{54.3} \color{red}{(+4.2)} & 54.6 & 77.3 & 416 & 247 & 1,771 & 2,679 & 0.8\\
\hline
GNNMatch \cite{papakis2020gcnnmatch} & 49.0 & 54.5 & 79.4 & 407 & 317 & 2,038 & 2,456 & 0.1\\
\textbf{TBooster+GNNMatch} & \textbf{53.4} \color{red}{(+4.4)} & 54.6 & 79.4 & 407 & 317 & 1.674 & 2,455 & 0.1\\
\hline
\end{tabular}
\end{center}
\end{table*}

\subsection{Datasets}
To evaluate the proposed tracklet booster, we conduct experiments on two widely used pedestrian tracking benchmark datasets, i.e., MOT17 \cite{milan2016mot16} and MOT20 \cite{dendorfer2020mot20}. The details of the datasets are described as follows.

\paragraph{\textbf{MOT17 dataset}} MOT17 is a widely used pedestrian tracking benchmark dataset. In total, there are $7$ training video sequences and $7$ testing video sequences. The benchmark also provides public deformable part models (DPM) \cite{felzenszwalb2010object}, Faster-RCNN \cite{ren2015faster} and scale dependent pooling (SDP) \cite{yang2016exploit} detections for both training and testing data. The number of tracks is 1,331 and the number of total frames is 11,235.

\paragraph{\textbf{MOT20 dataset}} MOT20 is a recently released pedestrian tracking dataset for crowded scenarios with plenty of occlusions. The average density is over 150. In total, there are $4$ training video sequences and $4$ testing video sequences. The dataset also provides public Faster R-CNN detections with ResNet101 \cite{he2016deep} as the backbone. The number of trajectories in the training data is 3,833 and the total number of frames is 13,410.

\subsection{Training Details}

The training details for the proposed tracklet booster are described as follows.

\paragraph{\textbf{Splitter}} We apply the simple and fast IOU tracker \cite{1517Bochinski2017} to the randomly jittered and horizontally flipped ground truth bounding boxes with variant overlapping thresholds to generate tracklets on the fly. Each tracklet may contain multiple ground truth object IDs since there are association errors. We store the ID-switch positions in the mask $\boldsymbol{m}^*$, which is treated as the ground truth splitting positions. The temporal window $T$ is set to be 65 frames. For tracklets that last shorter than 65 frames, padding with the starting and ending frames is used. A baseline re-id method \cite{Luo_2019_CVPR_Workshops} is adopted for extracting the input appearance feature with dimension 2048. 4 Normalized bounding box parameters are used as motion features. Thus, the input tracklet has a size of $2052 \times 65$.
We use Adam optimizer with an initial learning rate of 0.001. Cosine annealing learning rate scheduler is adopted with a maximum iteration of 60,000. 

\paragraph{\textbf{Connector}} For training Connector, we follow the similar data preparation as Splitter. Slightly different from the tracklets generated for training Splitter, we ensure that each tracklet only contains one object ID. 
Adam optimizer with initial learning rate 0.0001 is adopted. We use cosine annealing learning rate scheduler with maximum iteration 120,000. For hyper-parameters in the loss function, we set $\lambda=0.5$ and $\alpha=0.2$ in Eq.~(\ref{eq:loss_connector}).

\subsection{Inference Details}
\label{sec:inf_detail}

In the inference stage, we use tracklets generated from existing trackers as input. We process the tracklets with the temporal sliding window procedure with 50\% overlapping frames. Splitter is firstly conducted to predict the potential splitting positions. As defined in Section~\ref{sec:inference}, we set the splitting threshold $\delta_s=0.5$ for all experiments. Connector is also conducted in the sliding window manner. We set $\delta_t=64$ and $\delta_c=0.9$ for all following experiments. Currently, the proposed method is conducted offline. Since it is processed in a temporal sliding window manner, it can be modified to an online method, which will be implemented in future work.

\subsection{Evaluation Metrics}
We employ ID F1 measure (IDF1) \cite{ristani2016performance}, multiple object tracking accuracy (MOTA), mostly tracked targets (MT), mostly lost targets (ML), fragments (FM) and identity switches (IDS) \cite{milan2016mot16} as the evaluation metrics of the tracking performance, which are widely used in MOT. 

\subsection{Main Results}

We evaluate the performance of the proposed tracklet booster on MOT17 and MOT20 with several state-of-the-art (SOTA) methods.

\paragraph{\textbf{MOT17}} We test four SOTA methods, i.e., IOU \cite{1517Bochinski2017}, Tracktor \cite{tracktor_2019_ICCV}, MPNTrack \cite{braso2020learning} and CenterTrack \cite{zhou2020tracking} on MOT17, combined with the proposed tracklet booster method. 
Specifically, the IOU tracker uses intersection over union between bounding boxes across frames as the main clue for data association; Tracktor exploits the bounding box regression of an object detector to predict the position of an object in the next frame, given sequential input frames; MPNTrack exploits the classical network flow formulation of MOT to define a fully differentiable framework based on message passing networks; while CenterTrack applies detection to a pair of frames and associate the objects to the previous frame. The results are reported in Table~\ref{tab:MOT17}.
For each pair of comparisons, the original method is shown in the first row, while the method with the proposed TBooster (short for ``tracklet booster'') is shown in the second row. As expected, there are significant improvements on the IDF1 score, ranging from +0.6\% to +5.7\%. Since we do not tackle detection algorithms, the MOTA metrics roughly remain the same with TBooster added. 

\paragraph{\textbf{MOT20}} We also test four SOTA methods, i.e., Tracktor \cite{tracktor_2019_ICCV}, UnsupTrack \cite{karthik2020simple}, MOT20\_TBC \cite{ren2020tracking} and GCNNMatch \cite{papakis2020gcnnmatch} on MOT20, combined with the proposed tracklet booster method. 
Specifically, Tracktor is the same tracker as used in MOT17, UnsupTrack trains a ReID network to predict the generated labels using cross-entropy loss; MOT20\_TBC jointly models detection, counting,
and tracking of multiple targets as a network flow program,
which simultaneously finds the global optimal detections and
trajectories of multiple targets over the whole video; GCNNMatch uses graph convolutional neural network based feature extraction and end-to-end feature matching for object association. 
The results are reported in Table~\ref{tab:MOT20}. Similarly, there are also large improvements on the IDF1 compared with the original methods. This demonstrates the effectiveness of the proposed method on the association task in MOT.

\subsection{Qualitative Results}

\begin{figure*}[!t]
\begin{center}
\includegraphics[width=1.0\linewidth]{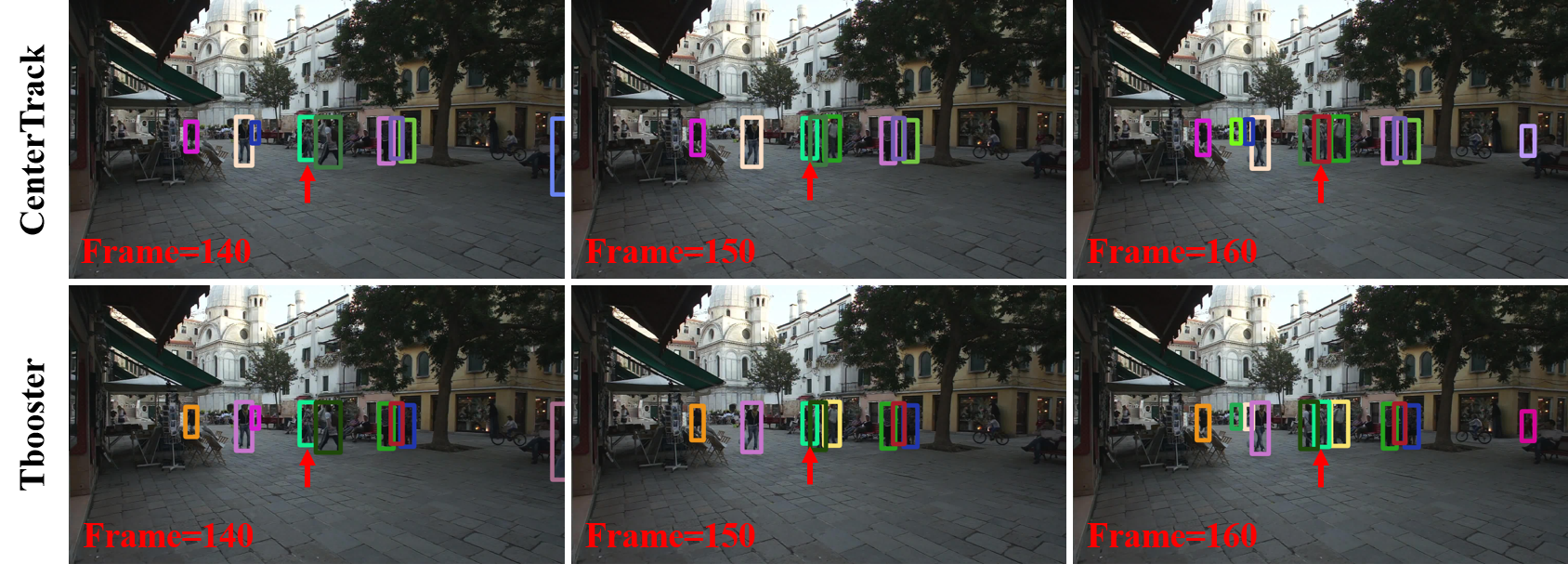}
\end{center}
   \caption{Qualitative comparison results between CenterTrack and TBooster. The frames are sampled from the sequence MOT17-01. Each color of the bounding box represents a distinct predicted object ID. The first row shows the result from CenterTrack and the second row is from TBooster. As the result from CenterTrack, the person with the pointed arrow changes ID after it reappears from occlusion at frame 160, while the association error is fixed in TBooster.}
\label{fig:vis}
\end{figure*}

To better visualize the improvement with TBooster, we show some qualitative comparison results between CenterTrack and TBooster in Figure~\ref{fig:vis}. The frames are sampled from the sequence MOT17-01. Each color of the bounding box represents a distinct predicted object ID. The first row shows the result from CenterTrack and the second row is from TBooster. As the result from CenterTrack, the person with the pointed arrow changes ID after it reappears from occlusion at frame 160. Thanks to Splitter and Connector modules, the association error is fixed in TBooster, as shown in the second row of the figure.

\begin{table}
\begin{center}
\caption{Effectiveness of Splitter and Connector.}
\label{tab:module}
\begin{tabular}{l|cc}
\hline
Method & IDF1 (\%) & MOTA (\%)\\
\hline
\hline
Original & 48.5 & 55.1 \\
w/. Connector & 52.2 \color{red}{(+3.7)} & 55.4 \\
w/. Splitter & 47.1 \color{blue}{(-1.4)} & 54.4 \\
w/. Splitter \& Connector & \textbf{54.6} \color{red}{(+6.1)} & 55.2 \\
\hline
\end{tabular}
\end{center}
\end{table}

\begin{table}
\begin{center}
\caption{Effect of different combinations of thresholds for Splitter and Connector.}
\label{tab:thres}
\begin{tabular}{l|cc}
\hline
Method & IDF1 (\%) & MOTA (\%) \\ 
\hline
\hline
Original & 48.5 & 55.1 \\ 
\hline
$\delta_s=0.5$, $\delta_c=0.5$ & 52.9 \color{red}{(+4.4)} & 54.9 \\ 
$\delta_s=0.5$, $\delta_c=0.7$ & 53.8 \color{red}{(+5.3)} & 55.1 \\ 
$\delta_s=0.5$, $\delta_c=0.9$ & \textbf{54.6} \color{red}{(+6.1)} & 55.2 \\ 
$\delta_s=0.5$, $\delta_c=1.1$ & 53.2 \color{red}{(+4.7)} & 55.2 \\ 
\hline
$\delta_s=0.7$, $\delta_c=0.5$ & 51.6 \color{red}{(+3.1)} & 55.2 \\ 
$\delta_s=0.7$, $\delta_c=0.7$ & 52.3 \color{red}{(+3.8)} & 55.3 \\ 
$\delta_s=0.7$, $\delta_c=0.9$ & 52.2 \color{red}{(+3.7)} & 55.3 \\ 
$\delta_s=0.7$, $\delta_c=1.1$ & 52.1 \color{red}{(+3.6)} & 55.3 \\ 
\hline
$\delta_s=0.9$, $\delta_c=0.5$ & 51.8 \color{red}{(+3.3)} & 55.3 \\ 
$\delta_s=0.9$, $\delta_c=0.7$ & 52.4 \color{red}{(+3.9)} & 55.4 \\ 
$\delta_s=0.9$, $\delta_c=0.9$ & 52.2 \color{red}{(+3.7)} & 55.4 \\ 
$\delta_s=0.9$, $\delta_c=1.1$ & 52.1 \color{red}{(+3.6)} & 55.4 \\ 
\hline
\end{tabular}
\end{center}
\end{table}



\subsection{Ablation Study}

For the ablation study, we use the MOT17-09 sequence as the validation set and the rest sequences as the training set. We adopt the IOU tracker as the baseline original tracking method.
\vspace*{+2mm}

\paragraph{\textbf{Study of the Effectiveness of the Splitter and Connector}} 
To test the effectiveness of Splitter and Connector modules, we compare three different settings, i.e., combining Connector without Splitter, combining Splitter without Connector, and combining both Splitter and Connector. The comparison results are shown in Table~\ref{tab:module}. As shown in the second row of the table, with a standalone Connector module, the IDF1 is boosted by 3.7\%. However, as shown in the third row, the IDF1 is decreased by 1.4\% with the standalone Splitter module. This means some tracks are divided into pieces, which negatively affects the IDF1 score. This is a common phenomenon, especially when occlusion happens. There is a high chance for tracklets to get split since the appearance feature changes rapidly. As shown in the last row of the table, with the combination of both Splitter and Connector, IDF1 is boosted by 6.1\%, which is a further boost of 2.4\% compared with standalone Connector module in the second row. This is because after the Splitter module, tracklets can get much purer IDs, which further helps Connector to increase the grouping accuracy. This demonstrates the effectiveness of both Splitter and Connector modules.

\paragraph{\textbf{Effect of Different Inference Thresholds}}
We also conduct experiments related to thresholds $\delta_s$ and $\delta_c$ for Splitter and Connector in the inference stage, respectively. We vary $\delta_s$ from 0.5 to 0.9 and $\delta_c$ from 0.5 to 1.1. We report the results in Table~\ref{tab:thres}. From the results, there is an over 3.0\% consistent improvement on the IDF1 score against the original tracking performance. This demonstrates the robustness of both Splitter and Connector modules.

\paragraph{\textbf{Effect of Adaptive Smoothing Strategy}}
To validate the effectiveness of the adaptive smoothing strategy in Splitter, we compare with the baseline method, where we use the ground truth splitting position $\boldsymbol{m}^*$ to guide the prediction $\boldsymbol{\hat m}$ directly using mean squared error without adaptive Gaussian smoothing strategy, i.e., the loss is set as $\mathcal{L}_s=||\boldsymbol{\hat m}-\boldsymbol{m}^*||^2$. 
In the evaluation, for both the baseline method and the proposed method, we select the local peaks from the predicted mask $\boldsymbol{\hat m}$ as the detected splitting positions and use corresponding peak values as the predicted confidences. Average precision (AP) is adopted as the evaluation metric for measure the splitting performance. We show the results in Table~\ref{tab:smooth}. Compared with the baseline method, there is a 9.4\% improvement on AP for the proposed approach, which shows the effectiveness of the adaptive Gaussian smoothing strategy.

\paragraph{\textbf{Effect of Multi-Head Self-Attention}}
To illustrate the influence of the multi-head self-attention mechanism, we conduct experiments with a variant number of attention heads in Connector. Specifically, we train Connector with four settings, i.e., with 1, 2, 4 and 8 heads, respectively. We assign $512/k$ channels to each head to ensure the total number of channels is fixed for all settings. We adopt the same training scheduler and keep other parts unchanged. The results are shown in Table~\ref{tab:head}. The best result is achieved when 4-head attention is employed and no further improvement with more attention heads used. Typically, more heads have the capability to deal with more complex situations. Meanwhile, the dimension of each head is reduced when we add more heads if the channel number is fixed. A trade-off should be made for both of the effects. Thus, the result also conforms to our expectations.

\begin{table}
\begin{center}
\caption{Comparison between adaptive smoothing and the baseline method in Splitter.}
\label{tab:smooth}
\begin{tabular}{lcc}
\hline
Method & Baseline & Adaptive Smoothing\\
\hline
\hline
Average Precision (\%) & 49.0 & \textbf{58.4} \\
\hline
\end{tabular}
\end{center}
\end{table}

\begin{table}
\begin{center}
\caption{Effect of varying number of heads in the self-attention.}
\label{tab:head}
\begin{tabular}{ccc}
\hline
\#Heads & IDF1 (\%) & MOTA (\%) \\ 
\hline
\hline
1 & 53.5 & 55.1 \\ 
2 & 53.9  & 55.2 \\ 
4 & \textbf{54.6}  & 55.2 \\ 
8 & 54.4 & 55.2 \\ 
\hline
\end{tabular}
\end{center}
\end{table}

\section{Conclusion}
In this paper, we propose a simple yet effective tracklet boosting approach, that can be easily combined with the existing trackers and boost the association performance. The tracklet booster has two main modules, i.e., Splitter and Connector. Splitter estimates the ID-switch positions and splits tracklets into small parts, while Connector merges multiple tracklets into clusters if they are from the same object. To stabilize the training of Splitter, a novel adaptive Gaussian smoothing strategy is proposed. To learn the discriminative tracklet embeddings, a multi-head self-attention mechanism is employed in Connector. We validate the tracklet booster on two widely used benchmark datasets, i.e., MOT17 and MOT20, and achieve significant improvement against several SOTA methods. Moreover, we also conduct sufficient experiments on several aspects of tracklet booster in the ablation study, which further proves the effectiveness of each module in the proposed method. Currently, we are still working on combining the graph neural networks with multi-head self-attention encoder for tracklet embeddings. This would be our potential direction for future work.

\bibliographystyle{IEEEtran}
\bibliography{sample}

\end{document}